\documentclass[runningheads]{llncs}

\usepackage{eccv}

\usepackage{eccvabbrv}

\usepackage{graphicx}
\usepackage{booktabs}
\usepackage{amsmath,amssymb,amsfonts}
\usepackage{multirow}
\usepackage{booktabs}
\usepackage{xcolor}
\usepackage{graphicx}

\usepackage[accsupp]{axessibility}  

\usepackage{hyperref}

\usepackage{orcidlink}

\begin{document}

\title{SPROUT: A Scalable Diffusion Foundation Model for Multi-Crop Plant Phenotyping} 
\titlerunning{SPROUT: A Scalable Diffusion Foundation Model}

\author{Shuai Xiang\inst{1} \and
James Burridge\inst{1} \and
Shouyang Liu\inst{2} \and
Hao Lu\inst{3} \and
Tokihiro Fukatsu\inst{4,5} \and
Yinqiang Zheng\inst{6} \and
Wei Guo\inst{1}\orcidlink{0000-0002-3017-5464}\thanks{Corresponding author.}}

\authorrunning{S. Xiang et al.}

\institute{Graduate School of Agricultural and Life Sciences,
The University of Tokyo, 1-1-1 Midori-cho, Nishitokyo,
Tokyo 188-0002, Japan\\
\email{\{xiang-shuai,burridge-j,guowei\}@g.ecc.u-tokyo.ac.jp}
\and
Engineering Research Center of Plant Phenotyping, Ministry of Education;
Jiangsu Collaborative Innovation Center for Modern Crop Production;
Academy for Advanced Interdisciplinary Studies,
Nanjing Agricultural University, Nanjing 210095, China\\
\email{shouyang.liu@njau.edu.cn}
\and
National Key Laboratory of Multispectral Information Intelligent
Processing Technology, School of Artificial Intelligence and Automation,
Huazhong University of Science and Technology, Wuhan 430074, China\\
\email{hlu@hust.edu.cn}
\and
Institute of Agricultural Machinery, NARO, 3-1-3 Kannondai,
Tsukuba, Ibaraki 305-8604, Japan\\
\email{fukatsu.tokihiro604@naro.go.jp}
\and
Institute of Life and Environmental Sciences, University of Tsukuba,
1-1-1 Tennodai, Tsukuba, Ibaraki 305-8572, Japan
\and
Next Generation Artificial Intelligence Research Center,
The University of Tokyo, Tokyo, Japan\\
\email{yqzheng@ai.u-tokyo.ac.jp}}

\maketitle

\begin{abstract}
  Image-based plant phenotyping depends on dense structural understanding of
  crops, yet pixel-level annotation remains expensive across species, organs,
  growth stages, and field conditions. General-purpose vision foundation
  models offer a natural route to label efficiency, but their web-scale
  pretraining objectives transfer weakly to agricultural imagery, where
  semantics are often determined by fine organ geometry inside repetitive,
  texture-dominated scenes. We introduce \textbf{SPROUT}, a diffusion
  foundation model for multi-crop plant phenotyping. SPROUT learns from
  2.6~million unlabeled open-field images (MCD-2.6M) using a pixel-space
  Diffusion Transformer, and selects transferable features with a label-free
  effective-rank criterion over denoising timesteps. This design shifts
  pretraining from crop-based invariance to structure-preserving denoising,
  making the representation better aligned with dense phenotyping tasks. We
  evaluate SPROUT across dense phenotyping tasks, including organ
  segmentation, crop--weed parsing, depth estimation, and counting.
  SPROUT consistently improves over strong web-pretrained baselines, with the
  largest gains on dense structural prediction, and shows favorable label and
  compute efficiency compared with general-purpose and crop-specific
  foundation models. The source code and MCD-2.6M dataset are publicly
  available on
  \href{https://github.com/UTokyo-FieldPhenomics-Lab/SPROUT}{GitHub}
  and
  \href{https://huggingface.co/datasets/XIANG-Shuai/MCD-2.6m}{Hugging Face},
  respectively.

  \keywords{Plant phenotyping \and
    Image-based phenotyping \and
    Foundation model \and
    Self-supervised learning \and
    Diffusion model \and
    Multi-crop generalisation}
\end{abstract}

\section{Introduction}
\label{sec:intro}

\begin{figure}[t]
  \centering
  \includegraphics[width=\linewidth]{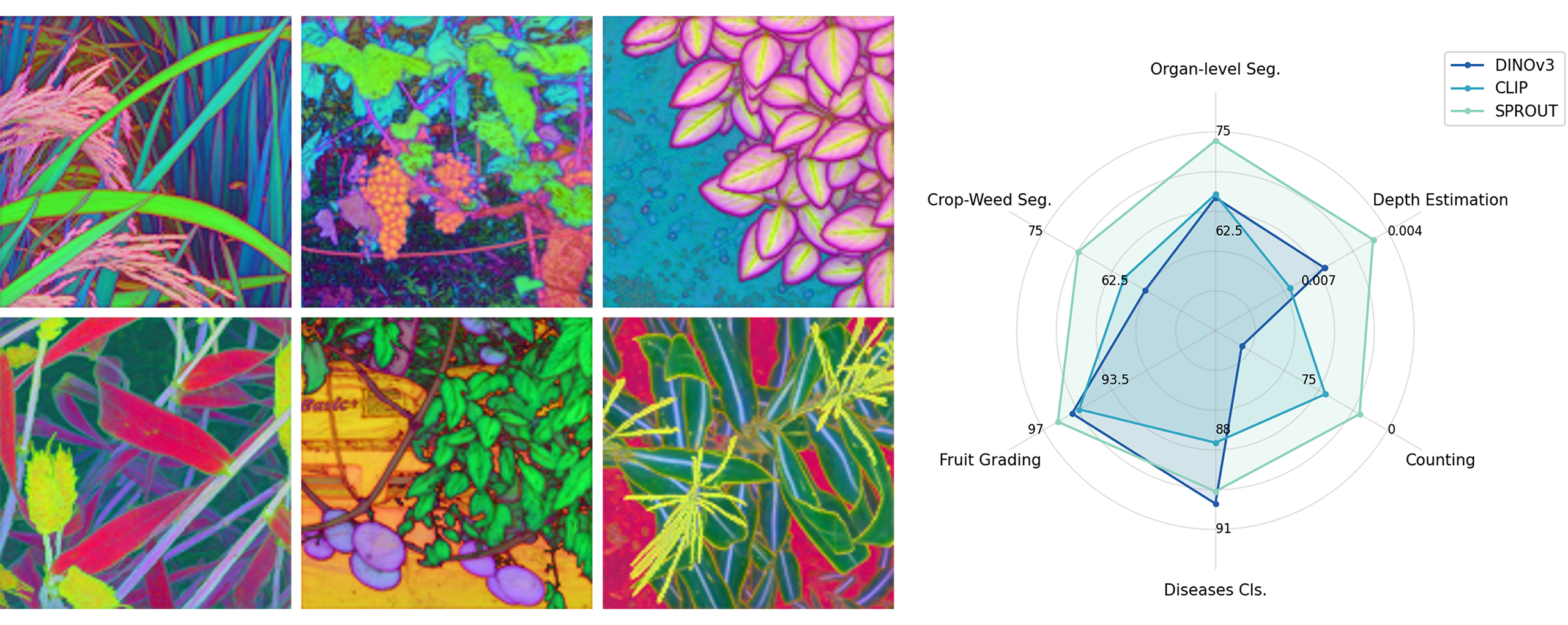}
  \caption{Left: PCA visualization of SPROUT feature maps. In the SPROUT embedding space, different plant organs exhibit clear semantic separation, while the same organ shares consistent semantics. SPROUT captures and understands the structural information of plants. Right: SPROUT's performance on agricultural vision tasks. Absolute Relative Error is the metric for depth estimation, Mean Square Error is the metric for counting, and Intersection over Union (IoU) is used for all other tasks. Metrics where lower values indicate better performance are inversely normalized. SPROUT significantly outperforms general-purpose VFMs across a wide range of tasks, particularly in structural understanding and dense prediction tasks.}
  \label{fig:Overview}
\end{figure}

Image-based plant phenotyping has become a cornerstone of modern crop
science, supporting trait discovery, breeding selection, precision
management, and high-throughput field experimentation
\cite{gwhd2021,gwfss,zhou2025global}. Across these applications, the
phenotypes that matter --- organ counts and sizes, plant architecture,
canopy structure, biomass distribution, weed--crop separation --- are
fundamentally \emph{dense} and \emph{structural}: they require pixel- or
organ-level understanding of plants in cluttered, texture-dominated field
scenes. Yet building such phenotyping pipelines still depends on bespoke,
densely annotated datasets per crop, per organ, per growth stage, and per
acquisition platform. Annotation is expensive, requires agronomic
expertise, and rarely transfers across crops; this annotation bottleneck,
not modelling capacity, is now the principal constraint on
high-throughput phenotyping at scale.

A natural way out of this bottleneck is foundation-model pretraining: a
single, label-free pretraining stage that produces image representations
transferable across crops and phenotyping tasks. General-purpose vision
foundation models (VFMs) trained on internet imagery --- CLIP
\cite{clip}, DINOv2 \cite{dinov2}, DINOv3 \cite{dinov3} --- have made
this approach standard in mainstream computer vision. Specialised
foundation models have followed in biomedicine
\cite{perez2025exploring,dai2025orochi}, remote sensing
\cite{tolan2024very}, and astronomy \cite{parker2024astroclip}, where
the visual statistics of the domain are sufficiently distinct from
internet imagery that domain-specific pretraining materially improves
downstream performance.

Plant phenotyping is one such domain, but foundation-model work here
remains early-stage and narrow. The handful of agricultural pretraining
efforts to date \cite{benito2025vision,shen2025weednet,han2025fomo4wheat}
are typically restricted to a single crop or a small task family, and
they inherit self-supervised objectives --- contrastive learning,
self-distillation, masked image modelling --- that were designed for
object-centric web imagery. Agricultural field images differ
fundamentally from that regime: they are texture-dominated, densely
populated by repeating plants with subtle organ-level differences, and
their semantically meaningful structure lives in fine spatial detail
rather than in object-versus-background contrast. The cropping- and
contrast-based view construction at the heart of contrastive and
self-distillation pretraining is therefore poorly matched to plant
imagery, and masked image modelling has so far either added substantial
architectural complexity or fallen short of state-of-the-art
self-supervised performance.

More concretely, the dominant self-supervised pretraining paradigms each
carry assumptions that fit poorly to field imagery. Contrastive learning
(e.g.\ SimCLR \cite{chen2020simple}) aligns differently augmented views of
the same image and pushes apart views of different images, and
self-distillation (e.g.\ DINO \cite{dino}) matches teacher and student
embeddings of paired views; both depend on random cropping to manufacture
positive and negative pairs. In agricultural scenes, two random crops of
the same field frequently contain visually similar plants from different
individuals, so the contrastive signal degrades. Masked image modelling
(MAE \cite{he2022masked}, BEiT \cite{beit}, iBOT \cite{ibot})
reconstructs missing patches from visible context within a single view and
sidesteps this issue, but existing formulations either add substantial
architectural complexity or fall short of state-of-the-art self-supervised
methods on downstream transfer.

We therefore revisit diffusion-based generative modelling as a pretraining
objective for plant phenotyping. Diffusion models learn to reverse a noise
corruption process by denoising at multiple noise levels, a label-free
objective that naturally rewards capturing fine-grained structure. Recent
Transformer-based diffusion architectures (DiT) \cite{peebles2023scalable}
exhibit strong scaling behaviour in generative modelling and have begun to
be repurposed as unsupervised representation learners. Crucially, the
denoising objective is defined within a single image and does not depend
on cropping-based view construction, making it well matched to the
texture-dominated, densely populated character of agricultural imagery.

Building on this observation, we present \textbf{SPROUT}, a
diffusion-pretrained foundation model designed for image-based plant
phenotyping. SPROUT abbreviates \textbf{S}calable \textbf{P}lant
\textbf{R}epresentation model via \textbf{O}pen-field
\textbf{U}nsupervised \textbf{T}raining. It introduces a VAE-free
Diffusion Transformer (UDiT)
operating directly in pixel space, which enables end-to-end optimisation
and avoids the inference latency and reconstruction bottleneck of latent
VAEs. We further address a practical obstacle to using diffusion features
for discriminative tasks --- the choice of denoising timestep --- by
proposing a label-free, training-free selection criterion based on the
effective rank of diffusion features, which consistently identifies
timesteps whose features transfer better to downstream phenotyping tasks.

To support pretraining at phenotyping-relevant scale we curate
\textbf{MCD-2.6M}, a multi-crop diffusion-pretraining corpus of
2.6~million open-field images distilled from an initial pool of
$\sim$4~million field- and web-collected samples spanning multiple sites,
seasons, and crop varieties; the corpus covers a broad range of plant
structures, growth stages, and field conditions.

Across phenotyping-aligned downstream evaluations --- organ-level
segmentation in apple, peach, pear, grape, wheat, and rice;
plant--weed separation across ten crop fields; canopy depth estimation
on sugar beet; and wheat-spike and soybean-pod counting --- SPROUT
consistently outperforms state-of-the-art web-pretrained VFMs on dense,
structure-dependent tasks, while requiring substantially less pretraining
compute than existing crop-specific foundation models.

Our main contributions are:
\begin{itemize}
    \item A diffusion-based pretraining framework tailored to plant
    phenotyping, combining a VAE-free pixel-space Diffusion Transformer
    with an effective-rank criterion for label-free timestep selection.
    \item A documented data-curation pipeline and the publicly released
    MCD-2.6M pretraining corpus of 2,604,450 images, with source-specific
    licensing and attribution information.
    \item \textbf{SPROUT}, a scalable, diffusion-pretrained foundation
    model for multi-crop, multi-task plant phenotyping.
    \item Phenotyping-aligned downstream evaluation showing that SPROUT
    improves organ- and plant-level segmentation, depth estimation, and
    yield-organ counting across diverse crops, while matching a
    specialist wheat foundation model with roughly one-twentieth the
    parameters and one-fortieth the pretraining compute, and reaching
    DINOv2-level accuracy with about $1/50$ of the labelled fine-tuning
    data.
\end{itemize}

\section{SPROUT}
\label{sec:SPROUT}

\subsection{Preliminaries}
\label{subsec:pl}

\subsubsection{Denoising Diffusion Models}
\label{subsubsec:dfm}
Diffusion-based generative models learn to reverse a gradual corruption
process from clean data to noise. They first perturb clean samples with a
noise schedule and then train a neural network to denoise the corrupted
samples.

Formally, given a clean sample $\mathbf{x}_0 \sim p(\mathbf{x}_0)$ and
Gaussian noise $\boldsymbol{\epsilon} \sim \mathcal{N}(\mathbf{0},
\mathbf{I})$, the forward diffusion process produces a noisy sample
$\mathbf{x}_t$ at timestep $t$:

\begin{equation} 
\label{eq:forward_z} 
\mathbf{x}_t = a(t)\mathbf{x}_0 + b(t)\boldsymbol{\epsilon} 
\end{equation}

where $a(t)$ and $b(t)$ are method-specific schedule functions governing the signal-to-noise ratio. As $t$ increases, $a(t) \to 0$ and $b(t) \to 1$, driving the distribution towards pure Gaussian noise.

To reverse this corruption, a denoising neural network
$D_{\boldsymbol{\theta}}$ is trained to regress a target variable
$r(\mathbf{x}_0,\boldsymbol{\epsilon},t)$ \cite{sun2025noise}. The
general optimization objective is:

\begin{equation}
  \label{eqn:opt_dfm}
  \begin{aligned}
  {\boldsymbol{\theta}}^* = \underset{\boldsymbol{\theta}}{\arg\min} \,
  & \mathbb{E}_{t \sim U(0, T)} \,
  \mathbb{E}_{\mathbf{x}_0 \sim p(\mathbf{x}_0)} \,
  \mathbb{E}_{\boldsymbol{\epsilon} \sim \mathcal{N}(\mathbf{0}, \mathbf{I})} & [\lambda(t)\| D_{\boldsymbol{\theta}}(\mathbf{x}_t, t) 
  - r(\mathbf{x}_0, \boldsymbol{\epsilon}, t) \|^2 ]
  \end{aligned}
\end{equation}

where $\lambda(t)$ is a weighting term. The regression target
$r(\mathbf{x}_0,\boldsymbol{\epsilon},t)$ is a combination of the clean
data and noise:
\begin{equation} 
\label{eq:r} 
r = c(t)\mathbf{x}_0 + d(t)\boldsymbol{\epsilon}
\end{equation}
The choice of $c(t)$ and $d(t)$ leads to three parameterizations:

\begin{enumerate}
    \item $\boldsymbol{\epsilon}$-prediction \cite{ddpm, ldm, peebles2023scalable}: $c(t) = 0$, $d(t) = 1$, and $r = \boldsymbol{\epsilon}$. The network estimates the added noise directly.
    \item $\mathbf{x}$-prediction \cite{ddpm, jit, delbracio2023inversion}: $c(t) = 1$, $d(t) = 0$, and $r = \mathbf{x}_0$. The network reconstructs the clean data from the noisy input $\mathbf{x}_t$.
    \item $\mathbf{v}$-prediction \cite{lipman2022flow}: $c(t) = -1$, $d(t) = 1$, and $r = \mathbf{v} = \boldsymbol{\epsilon} - \mathbf{x}_0$. This objective targets a velocity vector that combines data and noise.
\end{enumerate}

\subsection{Removing DiT's Dependency on VAE}
\label{sec:remove_vae}

Following LDM \cite{ldm}, DiT \cite{peebles2023scalable} encodes images from pixel space
to latent space with a VAE \cite{kingma2013auto} and trains the diffusion
model in that latent space. This design separates representation learning
from diffusion training: the VAE adds inference overhead, constrains
reconstruction quality, and cannot be optimized end-to-end with the
diffusion model. Several recent works therefore explore training DiT
directly in pixel space.

JiT \cite{jit} shows that pixel-space DiT training is most stable with
the $\mathbf{x}_0$ parameterization. Predicting either
$\boldsymbol{\epsilon}$ or $\mathbf{v}$ can be unstable because
$\mathbf{x}_0$ lies on a lower-dimensional image manifold, whereas
$\boldsymbol{\epsilon}$ and $\mathbf{v}$ span the high-dimensional pixel
space.

For representation learning, however, generative stability alone does
not determine the best training signal. Predicting
$\boldsymbol{\epsilon}$ encourages the model to represent high-rank
pixel-space variation, which can preserve fine-grained structure useful
for plant phenotyping. Our goal is therefore to use
$\boldsymbol{\epsilon}$-parameterized pixel-space diffusion while
retaining stable optimization.

Prior stabilization strategies take different routes. DSD \cite{dsd}
uses self-distillation to jointly train the pixel encoder and diffusion
model, while DeCo \cite{deco} augments DiT with a pixel decoder and
fine-grained $1\times1$ patching to preserve high-frequency information.
Despite their effectiveness, these methods still rely on
patchify/unpatchify operations to map between spatial layouts and token
channels. We instead remove the unpatchify step and replace
patchify/unpatchify with CNN-based downsampling and upsampling modules.
These CNN modules handle high-frequency image details, allowing the ViT
backbone to focus on semantic representation learning.

\subsection{Efficient Timestep Selection via Effective Rank}

When applying diffusion models to discriminative tasks, the timestep $t$ is a critical hyperparameter that determines the characteristics of the learned representations. Early timesteps (high noise levels) typically produce abstract but blurry features, while later timesteps (low noise levels) emphasize fine-grained structure and high-frequency textures, often at the cost of semantic abstraction. Selecting an appropriate timestep is therefore essential for obtaining high-quality representations.

Despite its importance, most existing work offers limited guidance on the choice of $t$. Existing strategies are typically either:

\begin{enumerate}
    \item exhaustive search, which evaluates the model on downstream datasets across many timesteps and is computationally costly;
    \item heuristic selection, which fixes a timestep based on empirical intuition and may not generalize across data distributions.
\end{enumerate}

To address these limitations, we propose a simple, training-free method for timestep selection based on the effective rank (erank) \cite{roy2007effective} of diffusion features.

Given a pretrained but not fine-tuned diffusion model, we extract feature matrices $\mathbf{F}_t \in \mathbb{R}^{N \times D}$ at candidate timesteps $t$, where $N$ denotes the number of samples and $D$ the feature dimension. Let ${\sigma_i(t)}_{i=1}^{Q}$ be the singular values of $\mathbf{F}_t$, with $Q = \min(N, D)$. We normalize the spectrum as $p_i = \frac{\sigma_i}{\sum_{j=1}^{Q} \sigma_j}$.

The effective rank of $\mathbf{F}_t$ is defined as
\begin{equation}
\label{eq:erank}
\mathrm{erank}(\mathbf{F}_t) = \exp\left( - \sum_{i=1}^{Q} p_i(t) \log p_i(t) \right)
\end{equation}

Intuitively, erank measures the intrinsic dimensionality of the representation by quantifying how uniformly information is distributed across singular directions. Prior work in self-supervised learning has shown that higher erank correlates strongly with richer and more transferable representations \cite{garrido2023rankme}.

This criterion is well matched to diffusion features because timestep
selection is a representation-selection problem rather than a
generation-quality problem. Maximizing erank favors timesteps whose
features distribute information across many active dimensions, providing
a label-free proxy for balancing semantic abstraction and fine-grained
spatial detail.

We leverage this property to select the timestep that yields the most informative diffusion features. Specifically, the optimal timestep is chosen by maximizing the effective rank:
\begin{equation}
\label{eq:tstar}
t^* = \arg\max_{t} (\mathrm{erank}(\mathbf{F}_t))
\end{equation}

This procedure requires no additional training or labeled data. By
iterating this calculation over candidate timesteps, the method selects a
timestep from the intrinsic dimensionality of the feature itself.

\subsection{Dataset Construction and Curation}
We curate MCD-2.6M from an initial pool of 4.3 million agricultural
images, combining long-term field acquisitions with web-collected crop
imagery. The curation pipeline removes low-quality, redundant, and
out-of-domain samples before pretraining:

The goal of this corpus is not only scale, but also phenotypic coverage.
Field imagery varies strongly with crop morphology, growth stage,
illumination, clutter, occlusion, and viewpoint. We therefore retain
visually and semantically diverse samples while removing images that
contribute little plant-structure information.

\paragraph{Visual quality filtering.}
We remove approximately 250K images with low visual fidelity, including
severely under- or over-exposed images and samples affected by heavy fog
or blur.

\paragraph{Feature-based filtering.}
To improve diversity and information density, we compute image embeddings
with an SSCD model and remove near-duplicates by cosine similarity. We
also extract patch-level DINOv2 \cite{dinov2} features and filter images
with low feature variance, such as clear skies or uniform backgrounds.
This stage removes 1.2M images.

\paragraph{Content filtering.}
Finally, we train a classifier to exclude non-biological or out-of-domain
content, such as machinery and infrastructure. This step removes
approximately 335K images and yields the final 2.6M-image pretraining
corpus.

\paragraph{Relation to downstream datasets.}
All downstream benchmark datasets other than GWFSS were entirely
excluded from MCD-2.6M. For GWFSS, MCD-2.6M contains other unlabeled
images from the broader collection, but the annotated benchmark crops
and their corresponding source images were excluded from pretraining.

\section{Experiments}
To thoroughly assess SPROUT's performance in agricultural vision, we assemble a large collection of publicly available agricultural datasets spanning diverse tasks and crop species, and compare it against representative baselines.

\subsection{Dense Representation Quality}
In this section, we assess the quality of the dense representations learned by SPROUT. We compare SPROUT with representative visual encoders trained on large-scale web datasets. These models fall into three categories: Masked Image Modeling (MAE, MSN), self-distillation (DINOv2, DINOv3), and contrastive learning (CLIP, SigLIP). Additionally, we include FOMO4Wheat, a DINOv2-based model specifically trained on wheat images, as a domain-specific baseline.

In the quantitative evaluation, all models except FOMO4Wheat use no auxiliary decoders. Instead, a single convolutional layer projects the final output features of each model. During fine-tuning, all model parameters are updated. For FOMO4Wheat, we adopt its Mask2Former decoder. Fine-tuning and inference are performed at a resolution of $256\times256$, and sliding-window inference is used for high-resolution images.

\begin{figure}[t]
  \centering
  \includegraphics[width=\linewidth]{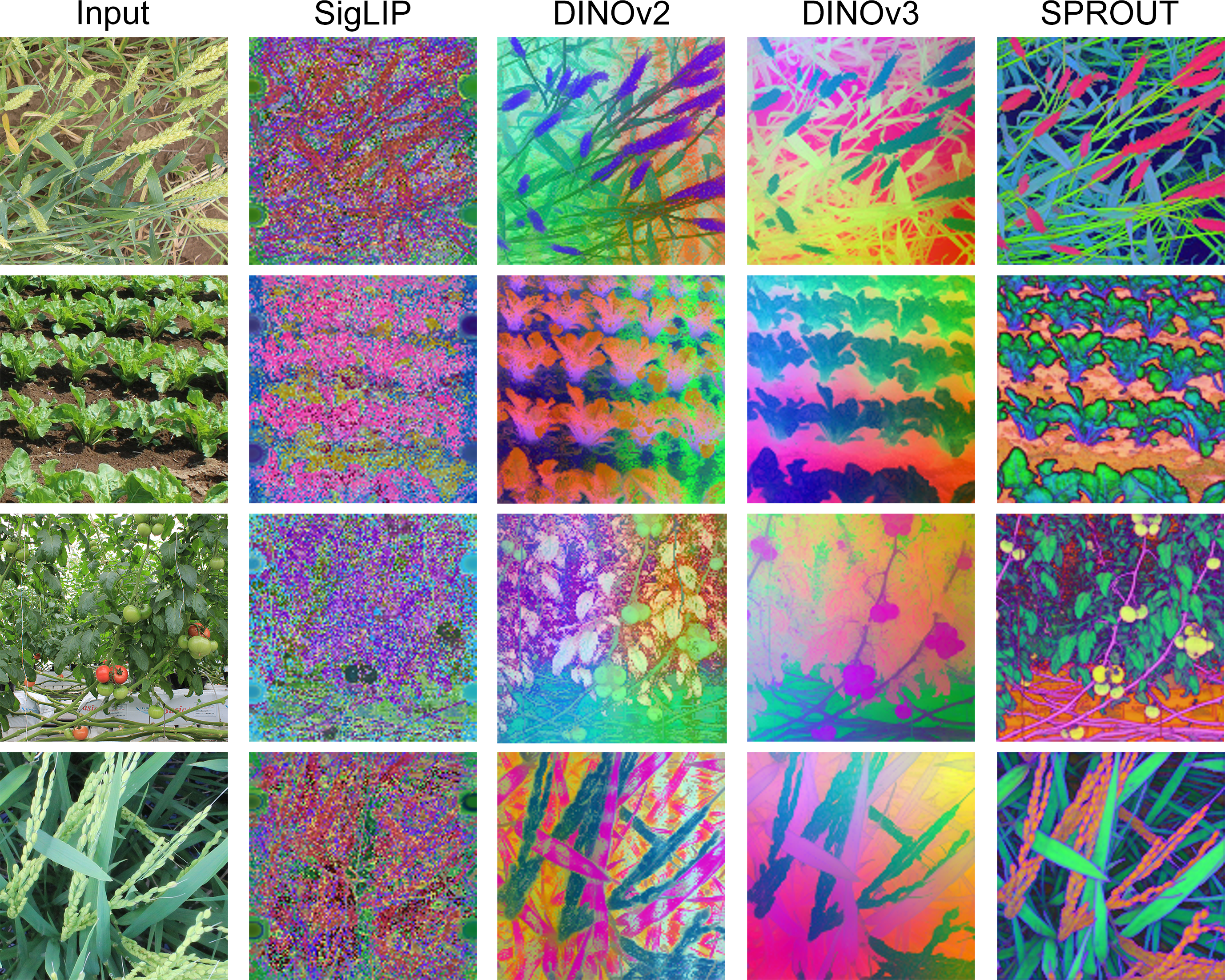}
  \caption{Comparison of dense features. We employ Principal Component Analysis (PCA) to reduce the dimensionality of the feature maps to three dimensions, projecting them into RGB space ($\mathbb{R}^{h \times w \times c} \to \mathbb{R}^{h \times w \times 3}$). Size of all feature maps is $128 \times 128$. Compared to prior methods, SPROUT yields clearer features with less noise and distinct semantics.}
  \label{fig:feature-map}
\end{figure}

\subsubsection{Qualitative Analysis of Dense Features}

We first conduct a qualitative analysis of the dense features produced by SPROUT. In the resulting RGB visualizations, similar colors indicate similar semantic information, while distinct colors represent semantically distinguishable regions. As illustrated in Fig.~\ref{fig:feature-map}, compared to visual encoders trained on web data, SPROUT maintains higher semantic consistency within the same organ type while differentiating distinct organs such as leaves, stems, and fruits. This indicates stronger sensitivity to plant structure.

This distinction is important for downstream phenotyping. Organ-level
traits are often defined by boundaries, thin structures, and repeated
parts rather than by image-level category cues. The visualization
therefore complements the quantitative segmentation results by showing
where the representation encodes plant structure before task-specific
fine-tuning.

\subsubsection{Organ-level Segmentation}
Organ-level segmentation, which aims to identify individual plant organs such as leaves, stems, spikes, flowers and fruits, plays a critical role in crop breeding. A core objective of breeding is to select optimal plant architectures with desired traits, including organ size, spatial layout, and biomass distribution. Since these traits are defined at the organ level, reliable quantification requires accurate organ-level understanding.

In this section, we evaluate the performance of different vision encoders in organ-level semantic segmentation. Table~\ref{t:organ-seg} reports the segmentation IoU of models pretrained with different strategies, including masked image modeling, self-distillation, and contrastive learning, together with SPROUT.

Across all crop and organ categories, SPROUT consistently achieves the best performance. This indicates that its representations capture plant structure prior rather than relying only on color or texture cues. This observation aligns with our qualitative results, where SPROUT's dense features show strong semantic consistency within the same organ class while remaining well separated across different organs. By enabling accurate and transferable organ-level analysis across diverse crops, SPROUT provides a solid basis for plant architecture selection and related downstream applications.

\begin{table*}[t]
    \caption{Organ-level semantic segmentation Intersection-over-Union (IoU) across multiple crops and organ categories. SPROUT consistently achieves the highest IoU across all crops and organs.}
    \label{t:organ-seg}
    \centering
    \setlength{\tabcolsep}{2pt}
    \resizebox{\textwidth}{!}{
      \begin{tabular}{lccccccccccccc}
        \toprule
        \multirow{2}{*}{Method} & \multirow{2}{*}{Pretraining dataset} & \multicolumn{1}{l}{\multirow{2}{*}{Model}} & \multicolumn{2}{c}{Apple} & Peach \cite{sun2021apple}  & Pear \cite{sun2021apple}   & Grape \cite{santos2020grape} & \multicolumn{3}{c}{Wheat \cite{gwfss}} & \multicolumn{3}{c}{Rice \cite{zhou2025global}} \\ \cmidrule(lr){4-5} \cmidrule(lr){6-6} \cmidrule(lr){7-7} \cmidrule(lr){8-8} \cmidrule(lr){9-11} \cmidrule(lr){12-14}
            &  & \multicolumn{1}{l}{}  & Flower \cite{sun2021apple}   & Fruit \cite{hani2020minneapple} & Flower   & Flower   & Fruit   & Spike & Stem  & Leaf  & Green Veg   & Senescent Veg  & Panicle  \\
        \midrule
        \textcolor{gray}{Self-Distillation} & & & & & & & & & & & & & \\
        DINOv2 \cite{dinov2} & LVD-142M & ViT-S-16 & 55.71 & 67.90 & 57.86 & 63.12 & 88.48 & 80.25 & 32.22 & 80.71 & 80.71 & 47.54 & 72.77 \\
        & & ViT-B-16 & 58.44 & 69.18 & 56.96 & 63.23 & 88.85 & 81.05 & 34.80 & 80.96 & 80.74 & 47.77 & 73.67 \\
        & & ViT-L-16 & 54.91 & 69.43 & 55.16 & 61.70 & 88.72 & 81.77 & 37.05 & 81.46 & 81.19 & 48.74 & 75.04 \\
        DINOv3 \cite{dinov3} & LVD-1689M & ViT-S-16 & 45.71 & 64.97 & 58.18 & 59.43 & 87.64 & 76.47 & 25.17 & 77.09 & 78.88 & 46.84 & 71.73 \\
        & & ViT-B-16 & 52.37 & 64.47 & 59.38 & 61.11 & 88.18 & 79.44 & 31.21 & 79.98 & 79.38 & 46.75 & 72.92 \\
        & & ViT-L-16 & 54.44 & 69.22 & 57.72 & 61.86 & 88.66 & 81.02 & 36.17 & 80.78 & 79.96 & 47.52 & 73.87 \\ 
        \midrule
        \textcolor{gray}{Masked Image Modeling} & & & & & & & & & & & & & \\
        MAE \cite{he2022masked} & ImageNet-1K & ViT-B-16 & 3.98 & 7.54 & 2.71 & 8.69 & 35.40 & 21.28 & 2.08 & 59.86 & 44.86 & 17.86 & 20.06 \\
        & & ViT-L-16 & 8.19 & 7.97 & 2.10 & 11.85 & 37.35 & 22.06 & 4.99 & 59.23 & 44.82 & 17.54 & 24.43 \\
        MSN \cite{msn} & ImageNet-1K & ViT-S-16 & 46.95 & 65.73 & 53.31 & 59.47 & 65.73 & 77.7 & 25.06 & 78.44 & 79.39 & 45.53 & 71.07 \\
        & & ViT-B-16 & 52.57 & 65.75 & 47.14 & 59.57 & 65.75 & 76.81 & 24.01 & 78.16 & 79.25 & 46.06 & 71.10 \\
        & & ViT-L-16 & 52.78 & 60.87 & 56.50 & 59.04 & 60.87 & 76.17 & 20.63 & 77.00 & 78.78 & 43.78 & 69.70 \\ 
        \midrule
        \textcolor{gray}{Contrastive Learning} & & & & & & & & & & & & & \\
        CLIP \cite{clip} & WIT-4M & ViT-B-16 & 55.39 & 67.08 & 59.61 & 59.72 & 88.10 & 78.3 & 26.32 & 78.88 & 79.40 & 44.96 & 70.05 \\
        & & ViT-L-14 & 57.71 & 70.20 & 57.22 & 62.40 & 89.00 & 81.3 & 32.43 & 81.51 & 82.01 & 50.06 & 74.77 \\
        SigLIP \cite{siglip} & WebLI-10B & ViT-B-16 & 54.86 & 66.57 & 57.73 & 59.93 & 87.78 & 76.77 & 25.2 & 78.52 & 78.99 & 45.51 & 71.31 \\
        & & ViT-L-16 & 55.32 & 66.99 & 56.79 & 60.71 & 87.81 & 77.49 & 26.36 & 79.02 & 79.48 & 45.74 & 71.92 \\ 
        \midrule
        \textcolor{gray}{Diffusion Denoising} & & & & & & & & & & & & & \\
        SPROUT (ours) & MCD-2.6M & UDiT-S & 56.61 & 71.92 & 34.34 & 51.19 & 89.72 & 82.43 & 38.23 & 83.52 & 86.80 & 51.18 & 78.60 \\
        & & UDiT-B & 59.26 & 73.10 & 51.71 & 71.01 & 90.06 & 84.37 & 44.31 & 85.00 & 86.82 & 53.12 & 80.06 \\
        & & UDiT-L & \textbf{65.92} & \textbf{74.28} & \textbf{62.09} & \textbf{73.03} & \textbf{90.63} & \textbf{85.54} & \textbf{51.58} & \textbf{86.44} & \textbf{87.11} & \textbf{55.29} & \textbf{81.15} \\ 
        \bottomrule
        \end{tabular}
    }
 \end{table*}

\subsubsection{Plant-level Segmentation}
Accurate plant-level segmentation is a prerequisite for precision weeding. Precision weeding enables targeted intervention on weeds while avoiding crops, thereby reducing herbicide usage and mitigating environmental impact. 

We evaluate different models on crop--weed segmentation across a variety
of crop fields. The Carrot dataset is from \cite{lameski2017weed}, Rice
is sourced from \cite{zhou2025global}, and the remaining datasets are
from CropAndWeed \cite{steininger2023cropandweed}. Several crops,
including Carrot, Pea, Potato, and Pumpkin, are absent from the
pretraining corpus. As shown in Table~\ref{t:weed-seg}, SPROUT-L achieves
the highest class-wise IoU in 18 of the 20 reported crop and weed
categories. CLIP-L obtains a higher IoU for potato-weed segmentation,
while DINOv2-L performs better for pumpkin-crop segmentation. Overall,
the results demonstrate broad cross-species transfer, including to crop
species not represented in the pretraining corpus.

\begin{table*}[t]
  \caption{Plant-level segmentation results on ten crop datasets. The
  table reports class-wise IoU for crop and weed segmentation. SPROUT
  achieves leading performance across most categories, indicating broad
  cross-species transfer.}
  \label{t:weed-seg}
  \centering
  \setlength{\tabcolsep}{2pt}
  \resizebox{\textwidth}{!}{
    \begin{tabular}{l*{20}{c}}
      \toprule
      Method    & \multicolumn{2}{l}{Bean}        & \multicolumn{2}{l}{Carrot}      & \multicolumn{2}{l}{Maize}       & \multicolumn{2}{l}{Pea} & \multicolumn{2}{l}{Potato}      & \multicolumn{2}{l}{Pumpkin}    & \multicolumn{2}{l}{Rice}      & \multicolumn{2}{l}{Soybean}     & \multicolumn{2}{l}{SugarBeet}   & \multicolumn{2}{l}{Sunflower}   \\
      \cmidrule(lr){2-3} \cmidrule(lr){4-5} \cmidrule(lr){6-7} \cmidrule(lr){8-9} \cmidrule(lr){10-11} \cmidrule(lr){12-13} \cmidrule(lr){14-15} \cmidrule(lr){16-17} \cmidrule(lr){18-19} \cmidrule(lr){20-21}
   & Crop  & Weed  & Crop  & Weed  & Crop  & Weed  & Crop  & Weed  & Crop  & Weed  & Crop & Weed  & Crop & Weed & Crop  & Weed  & Crop  & Weed  & Crop  & Weed  \\
    \midrule
      DINOv2-L  & 79.98 & 63.50 & 82.97 & 71.07 & 78.91 & 41.59 & 61.15 & 29.71 & 86.33 & 29.47 & \textbf{99.43} & 47.60 & 79.56 & 36.34 & 67.71 & 25.15 & 78.32 & 52.52 & 86.49 & 51.28 \\
      DINOv3-L  & 78.31 & 62.28 & 83.00 & 70.74 & 77.10 & 36.22 & 60.00 & 23.92 & 85.53 & 30.31 & 84.33 & 47.63 & 79.23 & 40.30 & 64.27 & 19.36 & 79.03 & 44.90 & 85.97 & 51.36 \\
      MSN-L     & 74.09 & 55.92 & 75.94 & 60.04 & 72.00 & 29.32 & 54.77 & 14.60 & 81.45 & 23.60 & 80.32 & 31.82 & 77.17 & 30.01 & 58.44 & 9.90  & 71.67 & 23.04 & 78.14 & 41.57 \\
      CLIP-L    & 80.96 & 64.67 & 84.20 & 73.02 & 79.88 & 42.22 & 61.53 & 30.72 & 87.41 & \textbf{34.46} & 86.63 & 51.03 & 80.70 & 39.54 & 68.33 & 28.87 & 80.96 & 51.84 & 86.65 & 51.35 \\
      SigLIP-L  & 77.49 & 59.23 & 81.46 & 68.93 & 74.43 & 35.71 & 58.23 & 19.90 & 84.70 & 25.80 & 81.82 & 40.15 & 77.87 & 36.68 & 62.75 & 20.81 & 77.75 & 43.08 & 84.46 & 49.09 \\
      \midrule
      SPROUT-S   & 84.40 & 64.47 & 85.69 & 74.85 & 87.70 & 51.44 & 72.22 & 32.49 & 91.87 & 25.62 & 92.54 & 59.02 & 85.77 & 37.24 & 79.47 & 27.14 & 87.08 & 49.63 & 89.83 & 53.60 \\
      SPROUT-B   & 85.42 & 68.46 & 85.58 & 74.53 & 87.69 & 51.23 & 70.54 & 33.41 & 91.72 & 28.17 & 92.57 & 60.61 & 86.48 & 40.71 & 80.15 & 31.14 & 85.67 & 50.30 & 89.08 & 51.39 \\
      SPROUT-L & \textbf{85.67} & \textbf{68.65} & \textbf{87.10} & \textbf{76.35} & \textbf{88.49} & \textbf{53.19} & \textbf{72.34} & \textbf{39.02} & \textbf{92.94} & 31.21 & 92.80 & \textbf{65.22} & \textbf{86.80} & \textbf{43.40} & \textbf{80.67} & \textbf{35.36} & \textbf{87.92} & \textbf{58.18} & \textbf{91.42} & \textbf{61.37} \\
      \bottomrule
    \end{tabular}
  }
\end{table*}

\subsubsection{Comparison with Single-crop Foundational Model}
We compare SPROUT with FOMO4Wheat \cite{han2025fomo4wheat}, a recently introduced agricultural foundation model for wheat. FOMO4Wheat uses the DINOv2 framework and is pretrained on 2.5 million wheat images. To ensure a fair comparison, we follow the same evaluation protocol as FOMO4Wheat, re-splitting the GWFSS dataset \cite{gwfss} into an 8:2 training and testing set rather than using the official competition split. Furthermore, all models were initialized randomly before pretraining to avoid reliance on external weights. 

As shown in Table~\ref{t:FOMO4Wheat}, our smallest variant, UDiT-S, achieves performance comparable to the largest FOMO4Wheat model (ViT-G) while utilizing approximately $1/22$ of the parameters. As model capacity increases, UDiT-L outperforms FOMO4Wheat, with particularly clear improvements on the most challenging category, \emph{Stem}, which requires fine-grained understanding of plant structure. Moreover, our approach requires a substantially lower computational cost. These results underscore both the efficiency and effectiveness of our pretraining strategy for agricultural data.

\begin{table*}[t]
  \caption{Comparison with the wheat foundational model FOMO4Wheat on wheat organ segmentation. One A100 hour refers to one hour of compute time using a single NVIDIA A100 GPU. Our SPROUT models obtain higher performance with markedly lower pretraining cost.}
  \label{t:FOMO4Wheat}
  \centering
  \setlength{\tabcolsep}{5pt}
  \resizebox{\textwidth}{!}{
    \begin{tabular}{lcccccc}
    \toprule
    Method     & Pretraining dataset & Model    & Params & Pretraining cost$\downarrow$ & Stem IoU$\uparrow$ & mIoU$\uparrow$  \\
    \midrule
    FOMO4Wheat & ImAg4Wheat-2.5M & ViT-G-16 & 1100 M & 9216 A100 Hours  & 46.85    & 74.63 \\
    \midrule
    SPROUT     & MCD-2.6M        & UDiT-S   & 51 M  & \textbf{245 A100 Hours}  & 48.84    & 74.77 \\
               &                 & UDiT-B   & 112 M  & 525 A100 Hours   & 52.88    & 76.46 \\
               &                 & UDiT-L   & 361 M   & 1440 A100 Hours   & \textbf{58.55}    & \textbf{78.38} \\
    \bottomrule
    \end{tabular}
  }
\end{table*}

\subsubsection{Depth Estimation}
Depth estimation provides dense geometric information about crops, which is essential for understanding canopy structure and monitoring growth dynamics. Depth cues are also a key component for enabling agricultural robots to perceive their environment.

We evaluate the proposed model on the sugar beet depth dataset \cite{chebrolu2017agricultural}, which provides real depth maps captured by Kinect v2. Quantitative results are reported in Table~\ref{t:deepth}. SPROUT achieves the lowest error across all evaluation metrics, indicating more accurate depth prediction.

\subsubsection{Counting}
In crop science, yield is often determined by the product of key organ counts. Counting yield organs such as spikes, pods, or grains from images is a direct and widely used approach for yield estimation. We evaluated our model's yield estimation performance using wheat \cite{gwhd2021} and soybean \cite{xiang2023yolo} datasets. 

As shown in Table~\ref{t:counting}, our method outperforms contemporary baselines, achieving the lowest MAE and MSE for both crops. These results indicate that SPROUT offers strong accuracy compared to standard object detection or density estimation methods.

\begin{table}[t]
  \centering
  \small
  \caption{Quantitative comparison of depth estimation performance on the Sugar Beet dataset.}
  \label{t:deepth}
  \begin{tabular}{lccc}
      \toprule
      Method   & AbsRel$\downarrow$ & MAE$\downarrow$ & RMSE$\downarrow$ \\
      \midrule
      DINOv2-L & 0.0060 & 0.0066 & 0.0103 \\
      DINOv3-L & 0.0062 & 0.0068 & 0.0110 \\
      CLIP-L   & 0.0074 & 0.0078 & 0.0139 \\
      \textbf{SPROUT-L} & \textbf{0.0045} & \textbf{0.0050} & \textbf{0.0071} \\
      \bottomrule
  \end{tabular}

  \caption{Counting performance on wheat spike and soybean pod datasets.}
  \label{t:counting}
  \begin{tabular}{lccc}
      \toprule
      Method & Supervision & MAE$\downarrow$  & MSE$\downarrow$   \\
      \midrule
      \textcolor{gray}{Wheat Spikes}  &  & & \\
      Faster R-CNN \cite{ren2016faster} & Box & 5.46 & 95.26 \\
      DETR \cite{detr} & Box & 8.41 & 262.44 \\
      YOLOv8 \cite{yolov8} & Box & 8.96 & 180.36 \\
      CSRNet \cite{csrnet} & Density Map & 5.88 & 66.10 \\
      TransCrowd \cite{liang2022transcrowd} & Density Map & 10.09 & 207.07 \\
      CSNet \cite{li2024csnet} & Count & 3.87 & 31.36 \\
      ResNet101 \cite{resnet} & Count & 7.54 & 94.94 \\
      DINOv3-L \cite{dinov3}  & Count & 6.44 & 126.97 \\
      MSN-L \cite{msn} & Count & 6.65 & 82.56 \\
      CLIP-L \cite{clip} & Count & 5.60 & 54.17 \\
      \textbf{SPROUT-L} & Count & \textbf{3.62} & \textbf{24.23} \\
      \midrule
      \textcolor{gray}{Soybean Pods} & & & \\
      ResNet101 \cite{resnet} & Count & 4.98 & 48.89 \\
      CLIP-L \cite{clip} & Count & 9.89 & 231.00 \\
      \textbf{SPROUT-L} & Count & \textbf{3.58} & \textbf{28.84} \\
      \bottomrule
  \end{tabular}
\end{table}

\subsection{Scalability and Efficiency}
In this section, we investigate the scalability of SPROUT pretraining and
its impact on downstream performance. Specifically, we fine-tune the
pretrained models on the official competition training split and evaluate
them on the validation set using 1-mIoU as the evaluation metric. All
fine-tuning hyperparameters are kept consistent across experiments.

\subsubsection{Scaling Model Size and Pretraining Compute}
As shown in Fig.~\ref{fig:scaling-compute-size}, when model size and dataset size are not limiting factors, we observe a clear power-law relationship between pretraining compute and downstream fine-tuning performance.

By simply scaling up the model size, fine-tuning performance on downstream tasks improves rapidly. This demonstrates that SPROUT exhibits strong scalability, making it well-suited for large-scale unsupervised learning. At equivalent computational budgets, larger models outperform smaller ones. However, extending the training duration for smaller models can also reduce the performance gap. This suggests that, for practical deployment, a smaller model with an extended training schedule might be preferable.

\subsubsection{Efficient Utilization of Labeled Data}
\begin{figure}[t]
  \centering
  \begin{minipage}[t]{0.55\linewidth}
    \centering
    \includegraphics[height=0.23\textheight,keepaspectratio]{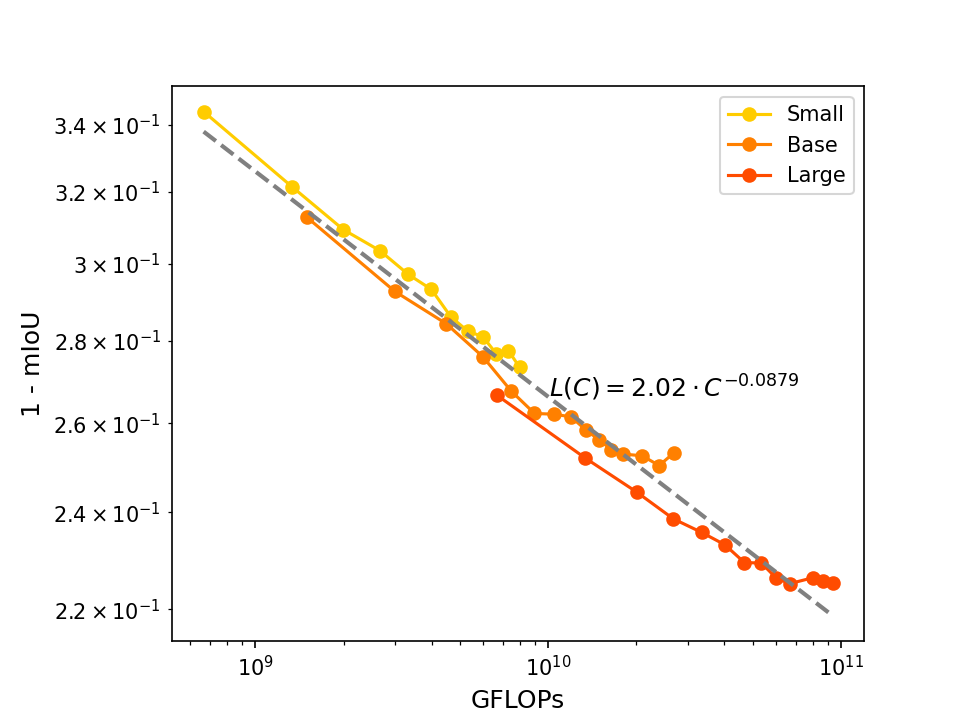}
    \caption{Scaling model size and pretraining compute. Downstream
    fine-tuning performance improves with increasing model size and
    pretraining compute.}
    \label{fig:scaling-compute-size}
  \end{minipage}
  \hfill
  \begin{minipage}[t]{0.40\linewidth}
    \centering
    \includegraphics[height=0.23\textheight,keepaspectratio]{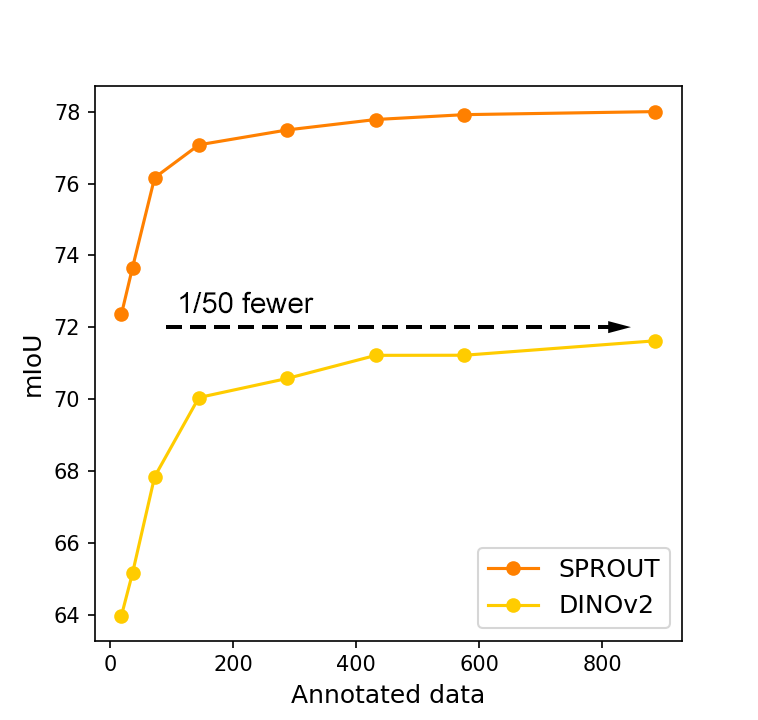}
    \caption{Efficient utilization of labeled data. SPROUT surpasses
    DINOv2 with 1/50 of the labeled data.}
    \label{fig:labeled-data}
  \end{minipage}
\end{figure}

We evaluate how effectively SPROUT leverages labeled data during
fine-tuning. As illustrated in Fig.~\ref{fig:labeled-data}, SPROUT
surpasses DINOv2 while using only 1/50 of the labeled data, demonstrating
its strong label efficiency and practical advantage in low-annotation
regimes.

\subsubsection{Scaling the Pretraining Dataset Size}
We further study how fine-tuning performance evolves as the volume of unlabeled pretraining data increases.

Fig.~\ref{fig:scaling-data} demonstrates that, as the unlabeled dataset is scaled up, downstream performance improves steadily when the dataset size is below 64K samples. However, once the dataset reaches approximately 64K, a clear saturation point emerges: further increasing the amount of homogeneous data yields diminishing returns. This observation suggests that future dataset construction should prioritize data diversity over simply increasing volume. In practice, more aggressive deduplication strategies can be adopted, and efforts should be made to collect data from diverse environments, crop types, and acquisition conditions.

\subsubsection{Estimation of the Optimal Number of Training Iterations}
\begin{figure}[t]
  \centering
  \includegraphics[width=0.95\linewidth]{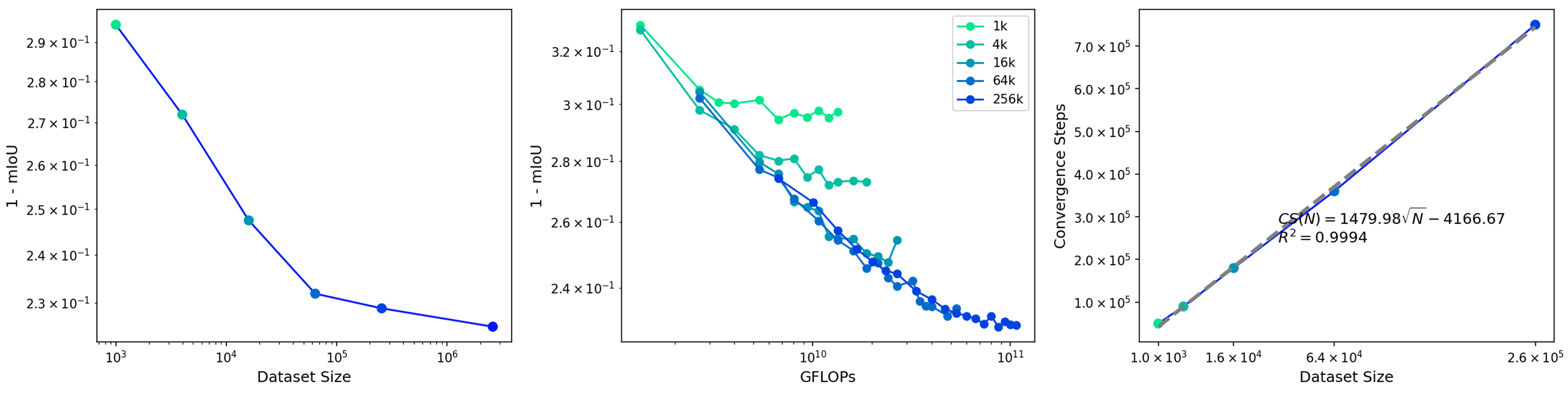}
  \caption{Scaling the pretraining dataset size. Left, downstream performance steadily improves as the unlabeled dataset scales up to $6.4\times10^4$ samples. After that, performance saturates, and adding more homogeneous data yields diminishing returns. Middle, model convergence under different dataset sizes. Right, convergence analysis of SPROUT-L across dataset scales. The required training steps scale linearly with the square root of dataset size, enabling estimation of the optimal number of training iterations.}
  \label{fig:scaling-data}
\end{figure}

To identify optimal training strategies, we explore the iterations required for SPROUT-L to reach convergence across different dataset sizes. As shown in the rightmost plot of Fig.~\ref{fig:scaling-data}, we find that the convergence training steps scale linearly with the square root of the dataset size. Therefore:

\noindent \textbf{For every four-fold increase in dataset size, the pretraining computational budget should be doubled.}

This empirical scaling rule offers practical guidance for allocating computing resources. For a new dataset, we can first extract a very small subset, calculate the number of steps required for convergence, and estimate the computational cost for convergence of the entire dataset.

\section{Conclusion}
\label{sec:conclusion}

We have presented \textbf{SPROUT}, a diffusion-pretrained foundation
model designed for image-based plant phenotyping. Pretrained by pixel-space
denoising on 2.6~million unlabeled open-field images spanning multiple
crops, growth stages, and sites, SPROUT yields representations that are
structurally faithful and transfer broadly across phenotyping-relevant
tasks: organ-level segmentation, plant--weed separation, canopy depth
estimation, and yield-organ counting. Across these tasks SPROUT
consistently outperforms web-pretrained vision foundation models on
dense, structure-dependent predictions, matches a specialist single-crop
foundation model with roughly one-twentieth the parameters and
one-fortieth the pretraining compute, and reaches DINOv2-level accuracy
with about $1/50$ of the labelled fine-tuning data. Important limitations
remain: the present model is RGB-only and single-view, was pretrained on
imagery dominated by temperate crops and growth conditions, and has not
yet been evaluated longitudinally for trait--yield inference. We see
extending diffusion pretraining to multispectral, multi-view, and temporal
field acquisitions as the natural next step.

\section*{Acknowledgements}
\label{sec:Ack}

This work was supported by the Japan Society for the Promotion of Science
(JSPS) KAKENHI under Grant Number 25H01110 and by the Japan Science and
Technology Agency (JST) under the AI-ENGAGE (Advancing Innovations for
Empowering NextGen AGriculturE) program through the project ``Bridging
Global Knowledge and Local Needs through AI for Enhanced Agricultural
Production, Sustainability and Resiliency (BRIDGE).'' The authors thank
the contributors of the public datasets used in the downstream evaluation
suite.

%
%
\bibliographystyle{splncs04}
\bibliography{ref}
\end{document}